\documentclass{article}

\usepackage[nonatbib,final,arx]{nips_2017}

\usepackage[utf8]{inputenc} 
\usepackage[T1]{fontenc}    
\usepackage{hyperref}       
\usepackage{url}            
\usepackage{booktabs}       
\usepackage{amsfonts}       
\usepackage{nicefrac}       
\usepackage{microtype}      
\usepackage[titletoc]{appendix}

\RequirePackage{graphicx}
\RequirePackage{amsmath}
\RequirePackage{amssymb}
\RequirePackage{amsthm}
\RequirePackage{amsfonts}
\RequirePackage{amsxtra}
\RequirePackage{subcaption}
\RequirePackage[
    backend=biber,
    bibencoding=utf8,
    sorting=none,
    sortcites=true,
    bibstyle=gost-numeric,
    citestyle=numeric-comp,
    autolang=other
]{biblatex}

\title{Learning to Generate Chairs with Generative Adversarial Nets}

\author{
  Evgeny Zamyatin\\
  Department of Computer Science\\
  ITMO University\\
  Saint-Petersburg, PA 199034\\
  \texttt{ezamyatin3@gmail.com}\\
   \And
  Andrey Filchenkov\\
  Department of Computer Science\\
  ITMO University\\
  Saint-Petersburg, PA 199034\\
  \texttt{aaafil@mail.ru} \\
}

\begin{document}

\maketitle

\begin{abstract}
  Generative adversarial networks (GANs) has gained tremendous popularity lately due to an ability to reinforce quality of its predictive model with generated objects and the quality of the generative model with and supervised feedback. GANs allow to synthesize images with a high degree of realism. However, the learning process of such models is a very complicated optimization problem and certain limitation for such models were found.
  It affects the choice of certain layers and nonlinearities when designing architectures. In particular, it does not allow to train convolutional GAN models with fully-connected hidden layers.
  In our work, we propose a modification of the previously described set of rules, as well as new approaches to designing architectures that will allow us to train more powerful GAN models. We show the effectiveness of our methods on the problem of synthesizing projections of 3D objects with the possibility of interpolation by class and view point.

\end{abstract}

\section{Introduction} 

Shortly after the great success achieved in tasks of image recognition and object detection by deep convolutional networks~\cite{krizhevsky2012imagenet,girshick2014rich,sermanet2013overfeat}, the problem of image synthesis has been recognized as an important one. Image synthesis finds application in the entertainment industry~\cite{reed2016generative,chen2017stylebank}, 3D modeling~\cite{girod2013principles}, medicine~\cite{onofrey2016mri} and many other areas~\cite{miyamotoapplication,chen2015using}.

In 2014, Alexey Dosovitsky et al.~\cite{DBLP:journals/corr/DosovitskiySB14} suggested a powerful generative model that was able to synthesize accurate images of 3D objects projections given a view point and its class, as well as suggest new images for previously unseen points of view and even interpolate between any two image classes, creating a line of morphed images, changing, for instance, from an armchair to a stool.

In the same year, Ian Goodfellow has introduced the revolutionary work Generative Adversarial Nets~\cite{2014arXiv1406.2661G}, in which he proposed a new concept of creating synthesizing networks. Then, in 2015, Alec Radford~\cite{DBLP:journals/corr/RadfordMC15} showed that it is possible to build deep convolutional models using GAN, called DCGAN, which are able to synthesize images with a high degree of realism. He also showed that the model doesn't just remember how certain images should look, but it is trained in general, and is capable of generating realistic images that were not represented in the dataset. But, unfortunately, it is very difficult to train convolutional GAN models, so the authors introduced a list of restrictions on the discriminator and generator architectures, following the rules of which it is possible to successfully train the models.

Following the rules of the list of restrictions from DCGAN, the concept of Generative Adversarial Nets can not be applied to the Dosovikiy generator: it is indicated in the restrictions that the fully-connected hidden layers should be removed wherever possible. But such layers are necessary, so that the model be able to build a powerful internal representation.

In our work, we propose the way to build more powerful GANs that can process images by adopting the best practices from generative and discriminative models. We show that this model is capable to synthesize more accurate images than any other model.

The rest of the paper is organized as follows.
Section~\ref{sec:related} contains a brief description of neural networks that allow to generate complex objects, especially images.
We present the approach in Section~\ref{sec:approach}.
In Section~\ref{sec:experiments}, we describe experiment setup and datasets we use.
Results of these experiments are presented in Section~\ref{sec:results}.
Section~\ref{sec:conclusion} conludes the paper.

You can find implementation here\footnote{https://github.com/EvgenyZamyatin/chair-gan-code}.

\section{Related Work}\label{sec:related} 

\subsection{Generative models} 

In paper~\cite{DBLP:journals/corr/DosovitskiySB14}, the authors addressed the problem of synthesizing the projections of 3D objects from a given view point and the class of the object. To solve it, a model of the neural network was developed, the architecture of which can be divided into two blocks:
\begin{enumerate}
  \item Folding the class and view point arguments to the internal feature vector using a block of the fully-connected hidden layers.
  \item Unfolding internal feature vector to the resulting image using a block of the the deconvolutional layers.
\end{enumerate}

\begin{figure}[h]
  \centering
  \includegraphics[width=.7\linewidth]{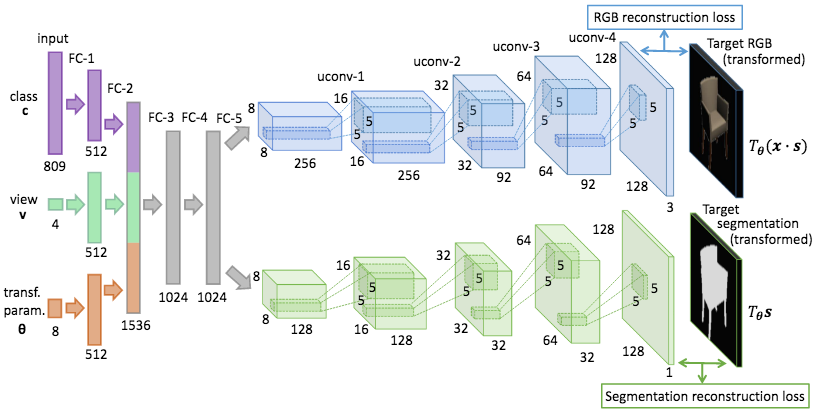}
  \caption{Dosovitskiy's generative model.}
  \label{fig:dosovitskiy-gen}
\end{figure}

The model was trained with the $l_2$-loss function. This model solves show good results in the task of generating images of 3D objects and show an ability to interpolate objects by class and angle. But the resulting images from this model are rather blurry.

In this work, the authors has used optical flow~\cite{Bro04a} algorithms after synthesizing for obtaining more sharp images.

In the papers \cite{DBLP:journals/corr/JohnsonAL16,DBLP:journals/corr/DumoulinSK16,DBLP:journals/corr/UlyanovVL16}, the authors presented models that were effective in the task of transferring the style of different pictures on the photo. Those models differed in that the image was generated not according to a parameter vector, but according to another photo containing required style.

\subsection{Normalization for GANs} 
The training of GANs is a very complicated optimization problem, the result of which depends heavily on the initialization of parameters, the learning rate and other structural parameters. The use of normalization in layers greatly helps in this situation.

Batch normalization~\cite{DBLP:journals/corr/IoffeS15} is a layer of deep neural network that normalizes the input data based on the statistics computed on the current batch. Thus, if we have a set of random variables distributed in accordance with the normal distribution as the input, then at the output we get a set of random variables distributed around zero with unit variance. Then we can shift and multiply the obtained distribution by the necessary quantities.

Instance normalization~\cite{DBLP:journals/corr/UlyanovVL16} is a normalization layer that is almost similar to the batch normalization, except that the statistic for normalization is evaluated in each instance of batch independently. This normalization is usually used when working with images.
It showed high efficiency on the problem of style transfer.

Weight normalization~\cite{DBLP:journals/corr/SalimansK16} is a normalization layer, which, as well as the previous one, was inspired by the batch normalization. Unlike the two previous types of normalization, this one does not use statistics counting at all. Instead, normalization occurs through the theoretical calculation of the mean and variance.
\begin{equation*}
\bar{x} \thicksim N(\mu_{x}, \sigma_{x}^{2})
\end{equation*}
\begin{equation*}
\bar{y} = \bar{x} \cdot W^{T}
\end{equation*}
\begin{equation*}
\bar{y} \thicksim N(\mu_{y}, \sigma_{y}^{2})
\end{equation*}
\begin{equation*}
\mu_{y, i} = \sum_{j} W_{i, j} \cdot \mu_{x}, \sigma_{y, i}^{2} = \sum_{j} W_{i, j}^{2} \cdot \sigma_{x}^{2}
\end{equation*}
\begin{equation*}
\frac{\bar{y}}{\sum_{j} W_{i, j}} \thicksim N(\mu_{x}, \sigma_{y}^{2})
\end{equation*}
\begin{equation*}
\frac{\bar{y}}{\sqrt{\sum_{j} W_{i, j}^{2}}} \thicksim N(\mu_{y}, \sigma_{x}^{2})
\end{equation*}
Thus, we can normalize the data without using the statistics of the batch.

\subsection{Generative Adversarial Nets} 

Generative Adversarial Networks is a new concept for training synthesizing models proposed by Ian Goodfellow et al.~\cite{2014arXiv1406.2661G}. This concept involve two functions: $G: Z \rightarrow X'$ representing a generator and $D: (X \cup X') \rightarrow S$ representing a discriminator, where $X$ denotes a set of elements from real distribution (that is dataset), $X'$ denotes a set of elements produced by the generator, $Z$ denotes some generation seed, usually random noise, and $S$ is a real number from $0$ to $1$ corresponding to whether an argument is received from $X$ or from $X'$.
$G$ and $D$ can be complex functions, such as neural networks. One network, which corresponds to $G$, tries how to emulate the distribution of $X$ from the noise $Z$. The second network, which corresponds to $D$, tries to learn how to separate the elements of the distribution $X$ and $G(Z)$. As the result, these components compete: $G$ tries to learn to fool $D$, and $D$ tries not to let $G$ do it.

Formally, we have two functions: $G(z, \theta_{g})$ that maps the random noise $z$ to a data distribution and which is parametrized by $\theta_{g}$, and $D(x, \theta_{d})$ that maps a data distribution in the interval $[0..1]$ and which is parameterized by $\theta_{d}$. $G$ is trained to minimize $\log(1 - D(G (z)))$, and $D$ is trained to maximize $\log(D(X))$. Thus, they play a min-max game with a value-function:
\begin{equation*}
  \min_{G}\max_{D} V (D, G) = \mathbb{E}_{x \sim p_{data}(x)} [\log D(x)] + \mathbb{E}_{z \sim p_{z}(z)} [\log(1 - D(G(z)))].
\end{equation*}

Deep Convolutional GANs (DCGANs) are a family of architectures, described in paper~\cite{DBLP:journals/corr/RadfordMC15}. Its authors introduced a list of limitations on the architecture of the discriminator and the generator and showed that such models were effective in the task of image synthesis. DCGANs are capable to produce highly realistic images. The restrictions imposed on the model consist of the following items:
\begin{enumerate}
  \item Replace each pooling layer with strided convolutions (discriminator) and fractional-strided convolutions (generator).
  \item Use batch normalization in both the generator and the discriminator.
  \item Remove fully-connected hidden layers for deeper architectures.
  \item Use ReLU activation in generator for all layers except for the output, which uses tanh.
  \item Use LeakyReLU activation in the discriminator for all layers.
\end{enumerate}

The authors of paper~\cite{DBLP:journals/corr/MirzaO14} proposed a method, by which the output of the generating network can be parameterized. Recall that previous models were trained in the unsupervised style from random noise $z$. In contrast, the authors propose to use a generator and a discriminator of the following form: $D = D(x | y)$ and $G = G(z | y)$. Thus, the discriminator says whether $x$ and $y$ are consistent, and the generator is trying to issue the appropriate $x$ for $z$ and $y$.
\begin{figure}[h]
  \centering
  \includegraphics[width=.8\linewidth]{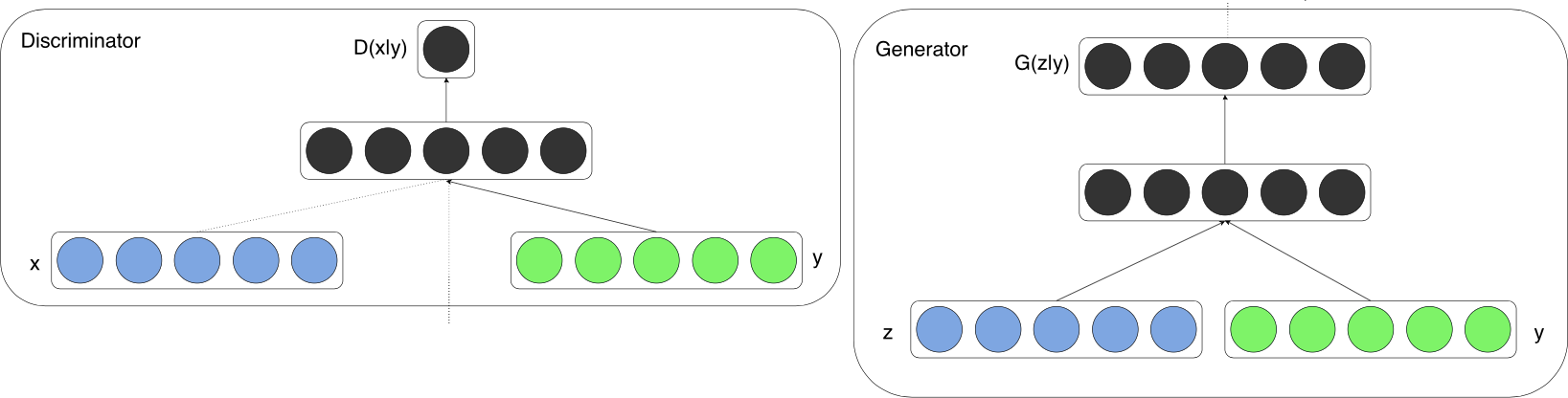}
  \caption{Conditional generative adversarial nets.}
  \label{fig:cgan}
\end{figure}

Various other models exist based on DCGAN architecture~\cite{DBLP:journals/corr/ChenDHSSA16,2016arXiv161009585O,DBLP:journals/corr/ZhuKSE16,DBLP:journals/corr/DosovitskiyB16,DBLP:journals/corr/White16a,2016arXiv160600704D}. In the~\cite{DBLP:journals/corr/SalimansGZCRC16,2017arXiv170107875A,2017arXiv170102386T} authors offered new methods for more stable learning of GAN models. In some works the authors use optical flow algorithms~\cite{Bro04a} to obtain more realistic images.


\section{Approach and Model Architecture}\label{sec:approach} 

In this section, we describe the model we suggest, namely the generator and discriminator, as well as the learning methods. There are distinguished cases for image generation: when only a single objects corresponds to a label (as in the well-known chairs dataset) and when there are lots of objects with a single label (as in ImageNet). The models used in these two tasks are quite different, but two our discriminators are very similar.

\subsection{Generator for absolutely conditional synthesis}\label{ssec:gen} 

In the first case, there is bijection between set of images and classes. The generator we use is based on the Dosovitsky's generator~\cite{DBLP:journals/corr/DosovitskiySB14}. As we previously described, the original model was designed to generate 3D objects, and coped quite well with this task. It was able to qualitatively interpolate objects by angle and class.
The architecture of the original generator is presented in Figure~\ref{fig:dosovitskiy-gen}.

However, it is impossible to simply use this architecture in GANs. Without changing the basics of this architecture, we can only add normalization layers and change the nonlinearities. We tried to follow the DCGAN pipeline where it was possible, so we decided to replace the nonlinearity at the network output with tanh and sigmoid. The use of batch normalization did not allow the successful training of the network, therefore, various normalizations were tried. We found that using instance normalization before the deconvolutional block and weight normalization after each deconvolution operation allows the network to learn much more efficiently and produce images of better quality.
So, summing up, we change this model in the following way:
\begin{enumerate}
  \item Before applying the image deconvolution, we use instance normalization~\cite{DBLP:journals/corr/UlyanovVL16}
  \item After each deconvolutional layer, we use weight normalization~\cite{DBLP:journals/corr/SalimansK16}
  \item In the end, we apply tanh for the rgb image and sigmoid to the mask.
\end{enumerate}

\begin{figure}[h]
  \centering
  \includegraphics[width=.6\linewidth]{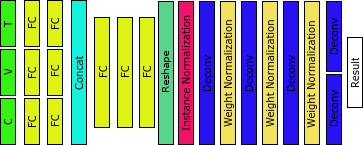}
  \caption{Generator architecture for absolutely conditional synthesis.}
  \label{fig:gen-schema}
\end{figure}

The architecture of the generator is present in figure~\ref{fig:gen-schema}.

\subsection{Generator for partially conditional synthesis}
\begin{figure}[h]
  \centering
  \includegraphics[width=.4\linewidth]{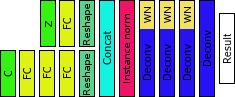}
  \caption{Generator architecture for partially conditional synthesis.}
  \label{fig:gen-schema-cifar}
\end{figure}

In the second case, several images has the same label, we slightly change the generator. We added random variable $\mathbf{z}$, which is responsible for the variety of images for a particular label as the input of the generator. We found that in the case of presence of $\mathbf{z}$, it is much more efficient to concatenate internal representations after reshaping. The generator architecture is presented in figure~\ref{fig:gen-schema-cifar}.

\subsection{Discriminator} 

As the architecture of the discriminator, we use the model of ``prover'' that is the binary classifier solving task of determination if input arguments correspond to the input object. Its goal is to check whether it is true that ``given object \textbf{$x$} corresponds to class $\mathbf{c}$''.
A similar approach is used by the authors of paper~\cite{DBLP:journals/corr/MirzaO14}. A schematic description of the model is presented in Figure~\ref{fig:cgan}.

In the discriminator, we use weight normalization after all convolutional layers instead of batch normalization, as in section~\ref{ssec:gen}. In all other aspects, we followed the DCGAN~\cite{DBLP:journals/corr/RadfordMC15} recommendations, except removing fully-connected hidden layers.
\begin{figure}[h]
  \centering
  \includegraphics[width=.4\linewidth]{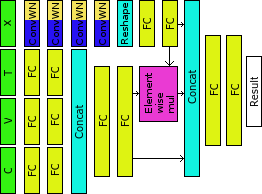}
  \caption{Discriminator architecture.}
  \label{fig:disc-schema}
\end{figure}

The architecture of the discriminator is present in figure~\ref{fig:disc-schema}. This architecture can be divided into three blocks:
\begin{enumerate}
  \item Information Hidden Vector Retrieval Block that is used for folding information vectors into the internal representation of the network.
  \item Image Hidden Vector Retrieval Block that is used for folding the image to an internal representation of the network.
  \item Hidden Vector Combination Block that is used for combining the resulting vectors with subsequent processing in order to obtain classification result.
\end{enumerate}
We describe these blocks in a more detailed way.

\textbf{Information Hidden Vector Retrieval Block.}
This part of the discriminator architecture is absolutely identical to the analogous block in the generative model. Each of information arguments is considered as an input for fully-connected layers. Output vectors are combined using concatenation with another subsequent application of the fully-connected layer. As a result, at the output of this block we get vector $\mathbf{x}_{info}$, which corresponds to the internal network representation of the information arguments.

\textbf{Image Hidden Vector Retrieval Block.}
This block includes a sequence of convolutional layers and application of the weight normalization after each convolution. After folding, we obtain a feature-vector of image. Then we process it with a fully-connected layer in order to obtain the resulting internal representation $\mathbf{x}_{img}$ of this image.

\textbf{Hidden Vector Combination Block.}
This block is designed to combine vectors that are provided by the blocks described above, namely internal representation of the information arguments $\mathbf{x}_{info}$ and internal representation of the image $\mathbf{x}_{img}$. The most straightforward way to combine these vectors is to use simple concatenation, as we did in Information Hidden Vector Retrieval Block.
However, the experiments showed that the use of such method does not allow the generator and discriminator to correctly learn and produce good images with correct correspondences to the information arguments.

We found the following approach to be effective in combining vectors $\mathbf{x}_{info}$ and $\mathbf{x}_{img}$: the vector of the internal representation of the arguments validity should be computed as
\begin{equation*}
  \mathbf{x}_{corr} = <\mathbf{x}_{info} \odot \mathbf{x}_{img}, \mathbf{x}_{info}, \mathbf{x}_{img}>,
\end{equation*}
where $\odot$ denotes the element-wise product, and $<>$ is the concatenation of its arguments. Vectors $\mathbf{x}_{info}$ and $\mathbf{x}_{img}$ must be of the same dimension.

Finally, a sequence of fully-connected layers with the sigmoid nonlinearity at the end is applied to the obtained vector {$\mathbf{x}_{corr}$}, the result of that is the degree of consistency of the given image and the given information arguments.

In the partially conditioned case the architecture of the discriminator remains the same, except for the difference in the number of conditional arguments. Also we use dropout in convolutional layers and $l_2$ regularization on the discriminator.

\subsection{Learning Algorithms} 

In the previous subsection, we described the generator that receives information vectors as the input and generates a suitable image and the discriminator the is being learned to distinguish if an information vector and an image are consistent.
The way how to train the generator is clear~--- it will be trained to ``fool'' the discriminator. That is, we will train the generator to produce such images for given arguments, that the discriminator will accept it. This leads to equation~\ref{eq:gen-loss}.
In order to create algorithms to learn the discriminator, we need to formalize its goal. Its output should be:
\begin{enumerate}
  \item $0$, if an unrealistic image is given (that is, obtained from the generator)
  \item $0$, if a realistic image is given (that is, an image from a dataset), but it does not correspond to a given set of information vectors
  \item $1$, if a realistic image is given and it corresponds to a given set of information vectors
\end{enumerate}
This description can be now used to suggest a loss functions that will be used for training.

In our task, the three values $(\mathbf{c}, \mathbf{v}, \mathbf{t})$ are the information vectors, which are the class of the object, the view point and the transformation.
The standard set of loss functions for the generator and discriminator of a GAN model is as follows:
\begin{equation}
V_{G}(\mathbf{c}, \mathbf{v}, \mathbf{t}) = BinaryCrossentropy(D(\mathbf{c}, \mathbf{v}, \mathbf{t}, G(\mathbf{c}, \mathbf{v}, \mathbf{t})), 1)
\label{eq:gen-loss}
\end{equation}
\begin{equation}
V_{D}^{gen}(\mathbf{c}, \mathbf{v}, \mathbf{t}) = BinaryCrossentropy(D(\mathbf{c}, \mathbf{v}, \mathbf{t}, G(\mathbf{c}, \mathbf{v}, \mathbf{t})), 0)
\label{eq:disc-loss-g}
\end{equation}
\begin{equation}
V_{D}^{real}(\mathbf{c}, \mathbf{v}, \mathbf{t}, x) = BinaryCrossentropy(D(\mathbf{c}, \mathbf{v}, \mathbf{t}, x), 1)
\label{eq:disc-loss-r}
\end{equation}
\begin{equation*}
BinaryCrossentropy(p, y) = -(y \cdot \log(p) + (1-y) \cdot \log(1-p)),
\end{equation*}
where $x$ is an image from the dataset. So, equations~\ref{eq:disc-loss-g} and~\ref{eq:disc-loss-r} corresponds to the first and the third item from discriminator behaviour description.

However, these functions are not enough, because we do not probed any information to the discriminator on non-matching images and their descriptions, which should be rejected. Therefore, the generator has the freedom to return images that do not correspond to the input arguments. This leads to the poor quality of the resulting images.

We handle this problem using \textit{negative sampling}. In addition to training the network on the correct data, we will specially add the data with incorrect correspondences. There are three vectors describing an image: $\mathbf{c}$ is the class label, $\mathbf{v}$ is the view point and $\mathbf{t}$ is the transformation vector. The simplest case is with $\mathbf{c}$: for each data set, we will generate one more exactly the same, replacing the class labels with random ones. We will get one more loss function for the discriminator:
\begin{equation*}
V_{D}^{neg_{c}}(\mathbf{c}, \mathbf{c}', \mathbf{v}, \mathbf{t}, x) = BinaryCrossentropy(D(\mathbf{c}', \mathbf{v}, \mathbf{t}, x), 0), \mathbf{c}' \neq \mathbf{c}
\end{equation*}

Since the class labels are a discrete value, they are not particularly problematic. $\mathbf{v}$ and $\mathbf{t}$ are continuous values, so we cannot simply replace them with others and claim that the correspondence in the data has become incorrect. Intuitively, the degree of correctness of the correspondence must be inversely proportional to the distance between the original (correct) vector and the random generated (wrong) one. So the loss function for $neg_{v}$ is as follows:
\begin{equation*}
V_{D}^{neg_{v}}(\mathbf{c}, \mathbf{v}, \mathbf{v}', \mathbf{t}, x) = \left\Vert \mathbf{v} - \mathbf{v}' \right\Vert^2 \cdot BinaryCrossentropy(D(\mathbf{c}, \mathbf{v}', \mathbf{t}, x), 0).
\end{equation*}
For $neg_{t}$, the situation is the same.

Now we have the resulting loss function for the discriminator, which will look like:
\begin{equation*}
V_{D} = \alpha \cdot V_{D}^{real} + \beta \cdot V_{D}^{gen} + \gamma_{c} \cdot V_{D}^{neg_{c}} + \gamma_{v} \cdot V_{D}^{neg_{v}} + \gamma_{t} \cdot V_{D}^{neg_{t}}
\end{equation*}

\section{Experiments}\label{sec:experiments}

\begin{figure}[h]
  \centering
  \includegraphics[width=.8\linewidth]{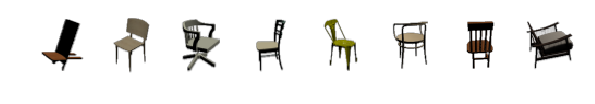}
  \caption{Dataset samples example.}
  \label{fig:dataset-example}
\end{figure}

In our work, we use three different dataset. The first of them is the well-known dataset that was used by Dosovitskiy et al.~\cite{DBLP:journals/corr/DosovitskiySB14}. This dataset contains a set of images of 3D chairs in various projections, as shown in figure~\ref{fig:dataset-example}. For each image, a label of the class $\mathbf{c}$ corresponding to the type of the chair presented in the image is known. Description of view point $\mathbf{v}$ is a set of four numbers corresponding to sinus and cosine of azimuth and altitude angles. Results on this dataset are presented in the body of the paper.
We also use augmentation for model training, identical to augmentation from the work of Dosovitsky et al. For each image, the transformation vector $\mathbf{t}$ was added.

The two other datasets are CelebFaces\footnote{\url{http://mmlab.ie.cuhk.edu.hk/projects/CelebA.html}} and Cifar10\footnote{\url{https://www.cs.toronto.edu/~kriz/cifar.html}}.
CelebFaces dataset is a set of images with faces and labels with it attributes.
Cifar10 is a set of images where each image is one from ten classes, such as airplane, car, horse, dog and so on.
In these datasets, the labels do not uniquely specify the images.
Results on these datasets are presented in the appendix.



We tested Dosovitskiy's generator learned with $l_2$-loss and bunch of different DCGAN models, including ACGAN~\cite{2016arXiv161009585O}, WGAN~\cite{2017arXiv170107875A}, improved GAN models~\cite{DBLP:journals/corr/SalimansGZCRC16} besides our model.
We train generator for 500 epochs. Composition of generator and discriminator we train 1000 epochs, but generator was trained every third time relative to discriminator. We use batches of size 16.

\section{Results and Discussion}\label{sec:results} 

\begin{figure}[h!]
\begin{subfigure}{\textwidth}
  \centering
  \includegraphics[width=.8\linewidth]{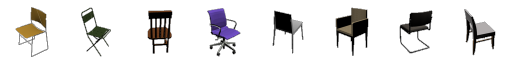}
  \caption{Chairs from dataset.}
\end{subfigure}
\begin{subfigure}{\textwidth}
  \centering
  \includegraphics[width=.8\linewidth]{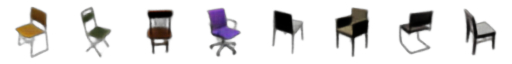}
  \caption{Samples from generator learned with l2-loss.(\textbf{Baseline})}
\end{subfigure}
\begin{subfigure}{\textwidth}
  \centering
  \includegraphics[width=.8\linewidth]{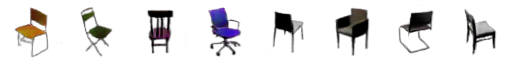}
  \caption{Samples from generator learned with gan-loss.(\textbf{Ours})}
\end{subfigure}
\caption{Random chairs.}
\label{fig:res-rand}
\end{figure}

\begin{figure}[h!]
\begin{subfigure}{\textwidth}
  \centering
  \includegraphics[width=.8\linewidth]{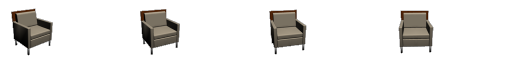}
  \caption{Chairs from dataset. Each second projection doesn't presented in dataset.}
\end{subfigure}
\begin{subfigure}{\textwidth}
  \centering
  \includegraphics[width=.8\linewidth]{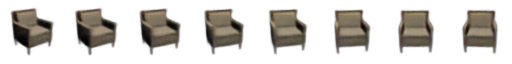}
  \caption{Samples from generator learned with l2-loss.(\textbf{Baseline})}
\end{subfigure}
\begin{subfigure}{\textwidth}
  \centering
  \includegraphics[width=.8\linewidth]{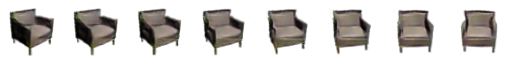}
  \caption{Samples from generator learned with gan-loss.(\textbf{Ours})}
\end{subfigure}
\caption{Rotation. Each second projection is not represented in the dataset.}
\label{fig:res-rot}
\end{figure}

\begin{figure}[h!]
\begin{subfigure}{\textwidth}
  \centering
  \includegraphics[width=.8\linewidth]{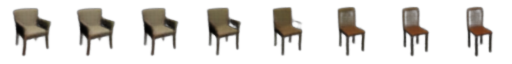}
  \caption{Samples from generator learned with l2-loss.(\textbf{Baseline})}
\end{subfigure}
\begin{subfigure}{\textwidth}
  \centering
  \includegraphics[width=.8\linewidth]{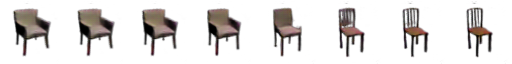}
  \caption{Samples from generator learned with gan-loss.(\textbf{Ours})}
\end{subfigure}
\caption{Class interpolation.}
\label{fig:res-inter}
\end{figure}

None of different DCGAN based models we tried coped with the task of accurately interpolating object viewing position. In paper~\cite{DBLP:journals/corr/ChenDHSSA16}, the authors evaluated their model on 3D chairs dataset, but their model produce images of less accuracy. Results of the chair generation, view point and class interpolation are presented in Figures~\ref{fig:res-rand}--\ref{fig:res-inter}.

According to the results of experiments on all three datasets, it is clear that the images obtained by the proposed generator trained with the discriminator, are more sharp. The generator also copes successfully with the problems of interpolation by angle and class.
More examples of network operation can be seen in the appendix.

It is worth paying attention to how well the generator that have never seen exact real images is able to be learned by the expense of a discriminator reproducing so accurately the smallest details of objects, correctly preserving view points.

\section{Conclusion}\label{sec:conclusion}

In our work, we combined the Generative Adversarial Nets capabilities with the power and flexibility of complex models, such as the Dosovitsky et al.'s generator~\cite{DBLP:journals/corr/DosovitskiySB14}. We presented new approaches for the design of generators and discriminators, which allow these models to be well trained in composition. We also showed the effectiveness of such models on the task of generating 3D object projections. The suggested architecture allows simple switching between who different cases: absolutely and partially conditional ones.

Based on this, we can formulate a modification of the rule set described in DCGAN~\cite{DBLP:journals/corr/RadfordMC15}, which allows to use fully-connected layers and getting more powerful GAN models:
\begin{enumerate}
  \item Use instance normalization~\cite{DBLP:journals/corr/UlyanovVL16} in the generator before starting the deconvolution of image
  \item In all convolutional and deconvolutional layers, use weight normalization~\cite{DBLP:journals/corr/SalimansK16}
  \item Use element-wise multiplication and negative sampling in conditional discriminator
\end{enumerate}

In addition to applying the GAN concept to create synthesis models capable of generating realistic images, GAN can also be used to improve quality and robustness of classifiers. Using the approaches described in our work, it is possible to build strong classifiers and train them using generators. In our work, we did not conduct such studies, but we plan to do it in the future.

\subsubsection*{Acknowledgments}
We’d like to thank Servers.com for donating a GPU server used in this work.

\printbibliography

\newpage
\begin{appendices}

\section{Additional images}

\begin{figure}[h!]
  \centering
  \includegraphics[height=\linewidth]{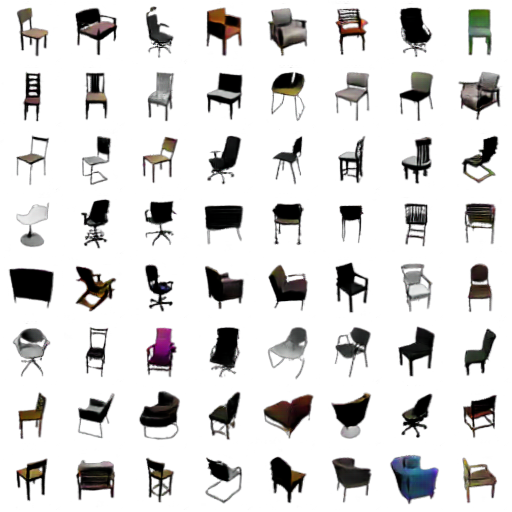}
  \caption{Set of random chairs in different projections.}
\end{figure}

\begin{figure}[h!]
  \centering
  \includegraphics[height=\linewidth]{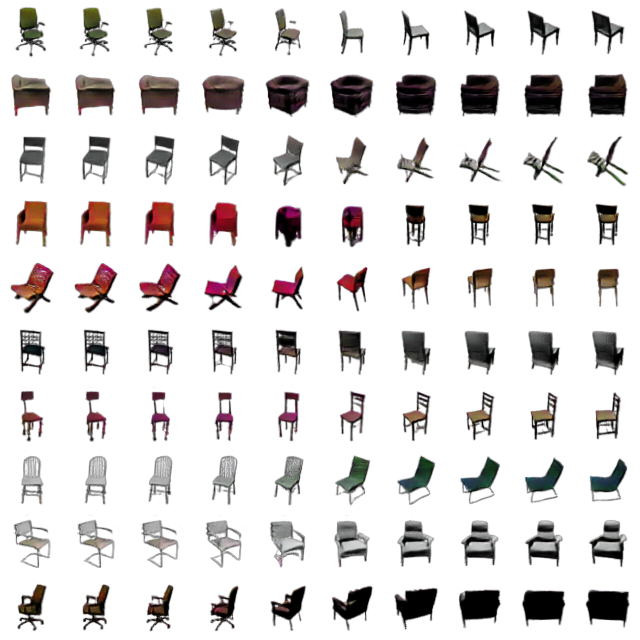}
  \caption{Set of random view point/class interpolations of chairs.}
\end{figure}

\begin{figure}[h!]
  \centering
  \includegraphics[width=\linewidth]{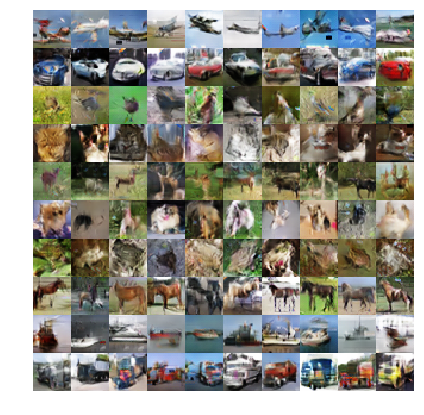}
  \caption{Generated images from Cifar10 dataset. The same class horizontally.}
\end{figure}

\begin{figure}[h!]
  \centering
  \includegraphics[width=\linewidth]{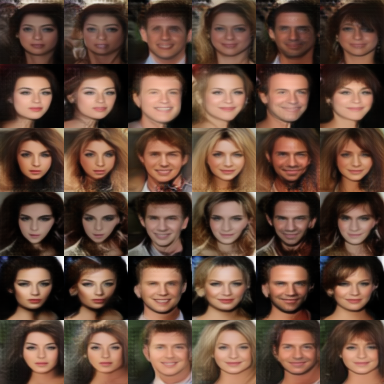}
  \caption{Generated images from CelebFaces dataset. Horizontally the same \textbf{$z$} var. Vertically the same \textbf{$c$} var.}
\end{figure}

\begin{figure}[h!]
  \centering
  \includegraphics[width=\linewidth]{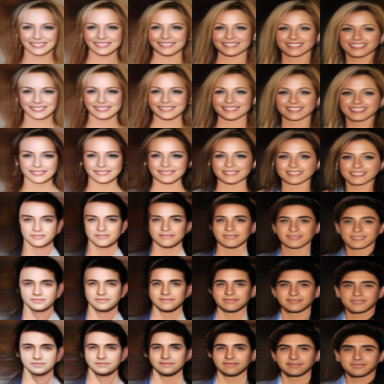}
  \caption{CelebFaces interpolation between two \textbf{$z$} vars(horizontally) and two \textbf{$c$} vars(vertically).}
\end{figure}

\end{appendices}

\end{document}